\documentclass[10pt,twocolumn,letterpaper]{article}

\usepackage{cvpr}
\usepackage{times}
\usepackage{epsfig}
\usepackage{graphicx}
\usepackage{amsmath}
\usepackage{amssymb}

\usepackage{mathrsfs}
\usepackage{float}
\usepackage{color}
\usepackage{ulem}
\usepackage{algorithm}
\usepackage{algorithmic}

\usepackage[breaklinks=true,bookmarks=false]{hyperref}

\cvprfinalcopy 

\def\cvprPaperID{****} 

\begin{document}

\title{Multi-object Tracking via End-to-end Tracklet Searching and  Ranking}

\author{Tao Hu \quad Lichao Huang \quad Han Shen\\
Horizon Robotics Inc\\
{\tt\small \{tao,hu, lichao.huang, han.shen\}@horizon.ai}}


\maketitle

\begin{abstract}
  Recent works in multiple object tracking use sequence model to calculate the similarity score between the detections and the previous tracklets. However, the forced exposure to ground-truth in the training stage leads to the training-inference discrepancy problem, i.e., exposure bias, where association error could accumulate in the inference and make the trajectories drift. In this paper, we propose a novel method for optimizing tracklet consistency, which directly takes the prediction errors into account by introducing an online, end-to-end tracklet search training process. Notably, our methods directly optimize the whole tracklet score instead of pairwise affinity. With sequence model  as appearance encoders of tracklet, our tracker achieves remarkable performance gain from conventional tracklet association baseline. Our methods have also achieved state-of-the-art in MOT15$\sim$17 challenge benchmarks using public detection and online settings.
\end{abstract}
\begin{figure}[ht]
\begin{center}
   \includegraphics[width=1.0\linewidth]{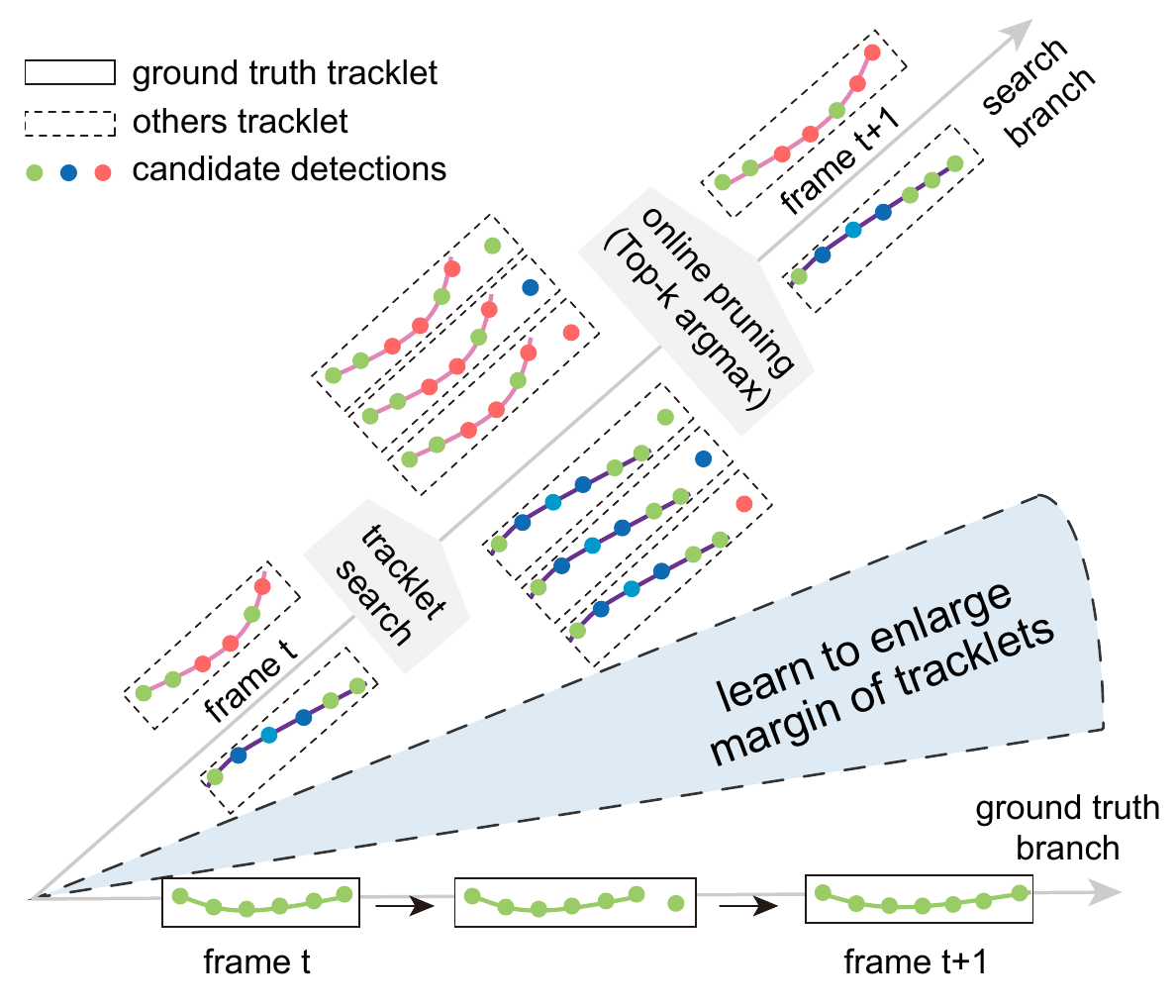}
\end{center}
\caption{The proposed methods of learning to score a tracklet base on online hypothesis search. Search branch: searched tracklet proposals from $t$ to $t+1$ in the online training. The tracklet proposals in time step $t+1$ are extended from previous time step. Only the Top$K$ score of tracklets are reserved for the loss calculations and the next step updates. Ground truth branch: The ground truth detection and trajectory is colored with green. The network learn to enlarge the margin between the ground truth tracklet and searched tracklet in search branch.}
\label{fig.tracklet_search}
\end{figure}

\section{Introduction}
\label{section:introduction}
Multiple object tracking (MOT), which aims to estimate the trajectories of several targets within a video sequence, is an essential but challenging task in computer vision~\cite{DBLP:journals/corr/abs-1907-12740}. From the machine learning perspective, tracking is the problem of consecutive sequence prediction and generation under given prerequisites. A common concern in tracking is how to prevent error accumulation when wrong prediction or association occurs, especially when the appearances of the neighboring individual objects are remarkably similar. With the impressive advances in deep learning based object detection algorithms~\cite{DBLP:journals/corr/HuangYDY15,DBLP:journals/pami/RenHG017,DBLP:conf/cvpr/YangCL16}, MOT community has strongly focused on the tracking-by-detection framework, which aims to link objects across frames correctly given detected bounding boxes~\cite{DBLP:journals/corr/abs-1907-12740}, a.k.a \textit{data association.} A popular choice for data association in multi-object tracking algorithm is pairwise-detection matching based on affinity model ~\cite{DBLP:conf/eccv/LiZG18a,DBLP:conf/icip/WojkeBP17,DBLP:conf/icip/BewleyGORU16,DBLP:conf/avss/BochinskiES17}. In these methods, detection results in adjacent frame are linked according to the affinity scores learned from appearance~\cite{DBLP:conf/eccv/LiZG18a,DBLP:conf/icip/WojkeBP17} or motion~\cite{DBLP:conf/icip/BewleyGORU16,DBLP:conf/avss/BochinskiES17} features. However, such methods only exploit the information of the current frame, but ignore the temporal cues from the previous frames in tracklets. As a results, such methods have limited capability to associate long-term consistent trajectories and are pruned to be trapped in local optimum by local matching. To address this issue, a few recent approaches~\cite{DBLP:conf/eccv/KimLR18,DBLP:conf/iccv/SadeghianAS17,DBLP:conf/aaai/MilanRD0S17,DBLP:conf/icmcs/MaYYZZJX18} build their affinity model on sequence model such as recurrent neural networks (RNNs~\cite{DBLP:journals/neco/HochreiterS97}). As a common practice, these methods usually force models to learn the affinity between tracklets and the candidate detections, with an objective to enlarge the affinity of correct matching and decrease the counterpart. Their works have demonstrated the effectiveness and potential of tracklet based methods to model higher-order information. However, these approaches suffer from two drawbacks. 
First of all, such methods which use recurrent neural networks to produce a tracklet representative feature for matching can somewhat be ill-posed. It is hard to force a sequence of targets with intra-variance to reach a consensus on appearance. Meanwhile, the final feature is inexplicable, not to mention in multi-cues engaged circumstances. 
Secondly, the previous tracklets such models trained on are assumed to be a pure detection sequence with the same ID, i.e. the ground truth tracklet. Nevertheless, it is not the case of inference, where wrong associations might occur at any time. This ideal assumption brings up a potential vulnerability that the model is trained on a different distribution from the test scenario, which can both diminish the discriminability and result in error accumulation during inference. This problem has also been emphasized in natural language processing~\cite{DBLP:conf/nips/BengioVJS15,DBLP:conf/nips/HeLXQ0L17,DBLP:journals/corr/RanzatoCAZ15,DBLP:conf/emnlp/WisemanR16} with a name of {\it exposure bias},  which occurs when a model is only exposed to the training data distribution. These earlier papers have also illustrated the importance of exposing the model to prediction data distribution.

In this work, we offer a possible solution to each of the two aforementioned issues. We propose a global score to measure the inner appearance consistency of tracklet, as opposed to measuring the affinity between the tracklet and target object. Notably, we optimize the whole tracklet with a margin loss. Besides, a novel algorithm has been established to simulate the prediction data distribution on training by introducing realistic discombobulated candidates to model. It helps to eliminate exposure bias problem to a great extent.

In summary, our algorithm has the following contributions:

1. We propose a tracklet scoring model based on margin loss and rank loss to quantify tracklet quality, which improves the tracklet consistency in data association.

2. We suggest a recurrent search-based optimization framework that remarkably exposes wrong associations to training. The training process follows a ``searching-learning-ranking-pruning" pipeline. It tackles the problem of exposure bias existing in sequence modeling that is neglected by previous MOT papers. 

3. Our method is validated on three benchmark datasets of MOT and achieves the state-of-the-art-results. We conduct extensive ablation studies and demonstrate the significant enhancement achieved by each component. Our code will be made publicly available after review.
\section{Related Work}
In this section, we give an overview of tracklet level tracking and approaches to reduce exposure bias related to our method.

\subsection{Tracklet Level Tracking Model}
With the improvement of the deep detector in recent years, tracking-by-detection paradigm~\cite{DBLP:phd/dnb/Chen15a} has become the most popular approach in MOT for its impressive performance. There are two mainstream categories methods of tracking-by-detection: tracklet level based tracking and pair-wise detection association approaches. Tracklet-level based tracking constructs an affinity model on the tracklet level and then uses it to associate the tracklet with detection or connect short tracklets. Pair-wise association methods establish an affinity model on the isolated detections, and then generate tracking results from the bottom up. The common concern of these two types of methods is to guarantee the consistency of the entire associated trajectories. Many previous approaches ~\cite{DBLP:journals/corr/abs-1904-04989,DBLP:conf/cvpr/SonBCH17,DBLP:conf/icip/WojkeBP17} have trained a binary classifier to determine the association between pairs of detections. However, such approaches are limited to modeling the very short-term correlation, i.e., two frames. It is difficult to model long-term temporal dependency and handle challenge scenarios such as appearance change. Some recent approaches build the association model on tracklet level to exploit the higher-order information~\cite{DBLP:conf/iccv/ChuOLWLY17,DBLP:conf/eccv/KimLR18,DBLP:conf/iccv/SadeghianAS17,DBLP:journals/corr/abs-1904-11489}. In these works, the tracklet representation comes from the fusion feature of individual detections through recurrent neural network~\cite{DBLP:conf/eccv/KimLR18,DBLP:conf/iccv/SadeghianAS17}, temporal attention~\cite{DBLP:conf/iccv/ChuOLWLY17} or relation model~\cite{DBLP:journals/corr/abs-1904-11489}. The positive results demonstrate that the long-term appearance information could be a favor to predict whether detection results belong to a given tracklet or not. However, the fused feature is not semantically explainable, because they are trying to find a most similar representative to the tailing target. On the other hand, such models are trained on the clip of the ground trajectories, which makes them vulnerable to exposure bias.
\subsection{Exposure bias in Tracking}
Exposure bias problem defines the phenomena that model is only exposed to the training data distribution (model distribution), instead of its own predictions (data distribution). The problem exists universally in machine learning related tasks, such as text summarization~\cite{DBLP:conf/conll/NallapatiZSGX16} and machine translation~\cite{DBLP:journals/corr/WuSCLNMKCGMKSJL16}, when prediction is made on a sequence of historical samples, all historical samples are fed with ground truth in training paradigm. Researchers in NLP have proposed their solutions~\cite{DBLP:conf/nips/BengioVJS15,DBLP:journals/corr/abs-1806-04936,DBLP:conf/naacl/TevetHSB19} to reduce the bias. Earlier, Bengio~\cite{DBLP:conf/nips/BengioVJS15} proposes a training schedule which make the model use itself's output as input with a increasing probability throughout training. Some researches attempt to avoid exposure bias by using non-conditional probability model. Semeniuta~\cite{DBLP:journals/corr/abs-1806-04936} proposes a “Reverse Language Model score” to evaluate the model’s generation performance. Tevet~\cite{DBLP:conf/naacl/TevetHSB19} uses Generative Adversarial Networks (GAN)~\cite{DBLP:journals/corr/Goodfellow17} to approximate the distribution of sequence. Such works motivate us to avoid exposure bias by applying the prediction data to model training.
Unfortunately, exposure bias have not received much attention from researchers in multiple object tracking community. The lastest work~\cite{DBLP:conf/cvpr/MaksaiF19} tries to eliminate exposure bias by designing an approximated IDF score loss, but it fails to balance the metrics of multi object tracking accuracy (MOTA) and IDF. Its result exhibits a tendency to drop less discriminative candidates according to the large number of false negative.

\begin{figure*}[ht]
\begin{center}
    \includegraphics[width=0.9\linewidth]{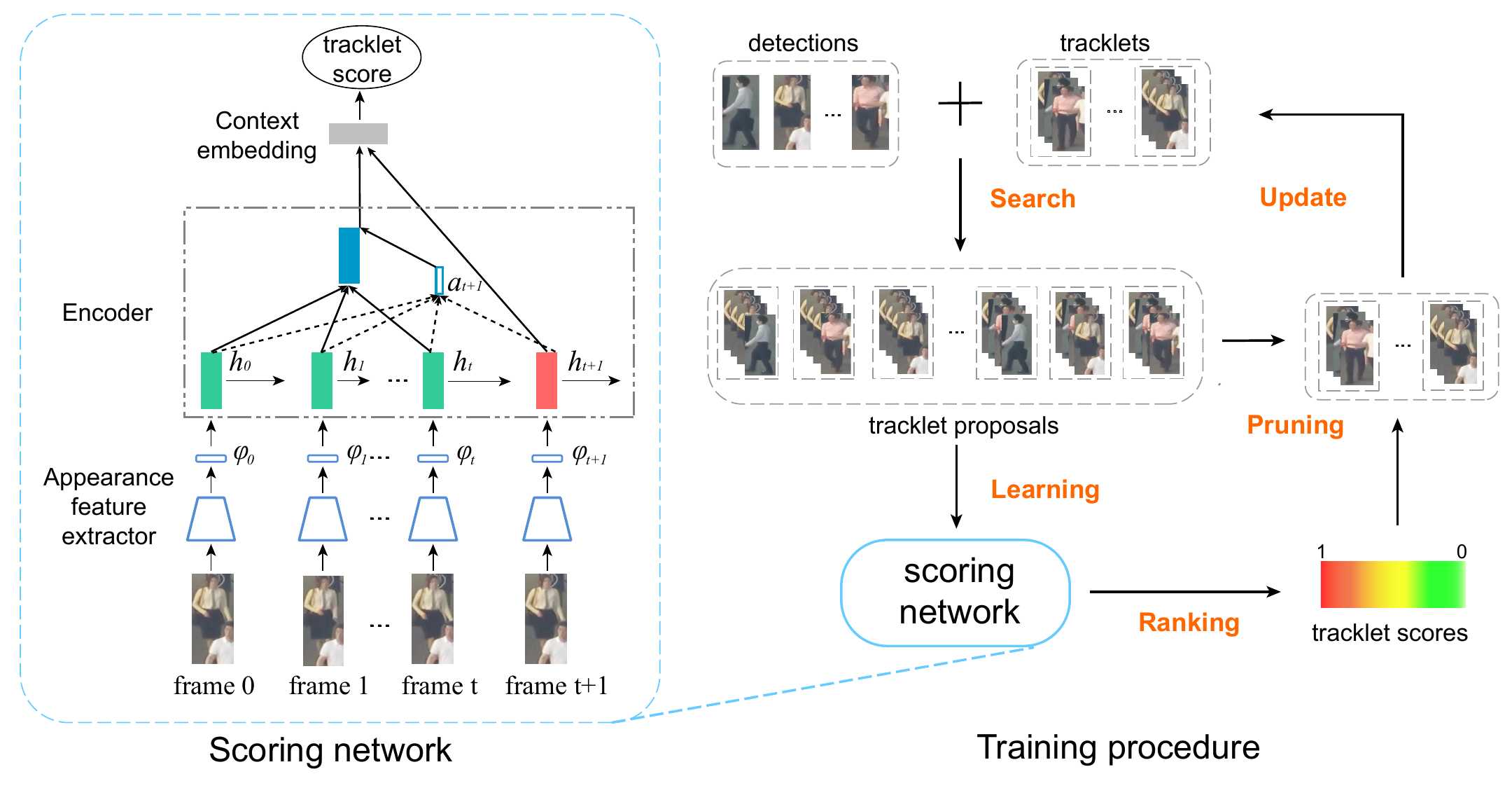}
\end{center}
   \caption{The overall of our proposed network architecture in training. The appearance feature of each detection are extracted with CNN (the blue hollow trapezoid), and the appearance embedding of tracklet are obtained through encoder (the gray dash rectangle) network which trained by online hypothesis tracklet searching with margin loss and rank loss. The tracklet score are used to represents the tracklet consistency, and then yield a set of hypothesis tracklets frame-by-frame with online search.}
\label{fig.overall.arch}
\end{figure*}

\section{Methods}
In tracking-by-detection paradigm, MOT is composed of two stages: detecting objects at each frame ({\it detection}) and assigning a track ID to each detection results across frames ({\it association}). A common practice to assign the IDs is usually according to the affinities between tracklets/detections. In this section, we  give an overall architecture description of our proposed framework with tracking-by detection paradigm firstly and then expand the details in the sub-sections.

We propose a new framework to directly optimize the tracklet score with margin loss by tracklet online searching. The framework is composed of an iterative searching, learning, ranking and pruning process as shown in Figure \ref{fig.overall.arch}. After obtaining a model, we adopt both the online Hungarian algorithm~\cite{munkres1957algorithms} and near online algorithm - multiple hypothesis tracking (MHT)~\cite{DBLP:conf/iccv/KimLCR15} in inference for an comprehensive illustration of its effectiveness. Finally we run this assignment process frame-by-frame and yield a set of target trajectories over time.

\subsection{Tracklet Level Optimization}
\label{section:methods:tracklet level optimization}
As illustrated in Section~\ref{section:introduction}, to address the issue of exposure bias and RNN based affinity model, we propose a novel framework to optimize the tracklet by ``searching-learning-ranking-pruning'' paradigm. The core parts include how we perform score learning, and how we conduct search-based tracklet optimization.
\noindent
\noindent
\textbf{Learning to Score.}
Let $T_{i}$ denotes the trajectory of object $i$ in a video, and it consists of $D_i$ detection patches $T_{i} = \{b_i^t\}_{t=t_0}^{t_{0}+D_i}$. Given an already obtained tracklets at time step $t$ as $T_i^{t}$, and a series of new observation detection results $\{b_1^{t+1}, b_2^{t+1}, ..., b_i^{t+1}, ..., b_j^{t+1}\}$. We present a tracklet search based method to optimize the consistency score of the extended tracklet $\left(T_i^{t}, b_j^{t+1}\right)$ (as described in Figure \ref{fig.tracklet_search}). 
Our goal is to find a scoring function which is to favor the consistency between the training and inference stage. Assuming that the scoring function $f_s\left( T \right)$ implemented with deep network is given, we firstly explore how to learn it by end-to-end training, while leave the network design to Section~\ref{section:methods:online appearance encoder}.

For a tracked object, assume that we have a set of predicted candidate tracklets  $\mathcal{T}_{t} = \{ \hat{T_1^t},  \hat{T_2^t}, ..., \hat{T}_i^t\}$ at time step $t$, where $\hat{T_i^t}=\{\hat{T}_{i}^{t-1}, b_j^t\}$. Tracking can be understood as to maximize the score of tracklet $T_{gt}^t$ which is consistent with ground truth tracklet and minimize the score of wrong connected tracklets $\mathcal{T}_{t} \backslash \{T_{gt}^t\}$. Rather than placing a hard constraint on the value of tracklet score, we prefer to leave some space for intra-instance variance, but punish the wrong associations if their scores may lead to an ambiguity. We then define a margin loss that constraint the score of ground truth tracklet to exceed the score of incorrect tracklet by a margin $\alpha$:
\begin{equation}
\begin{aligned}
L^{margin}_{t} &= \sum_{\hat{T}_i^t \in \{\mathcal{T}_{t} \backslash T_{gt}^t\}} max \left(0, \alpha - {\rm Sigmoid}\left( f_s\left(T_{gt}^t\right)\right) + \right. \\
& \left. {\rm Sigmoid} \left(f_s\left(\hat{T}_i^t\right)\right) \right)
\end{aligned}
\label{equation:margin_loss}
\end{equation}

The margin loss tries to distinguish the ground-truth tracklets from the predicted candidates, while can not quantify the differences between the candidates. The candidate with lower identity switch (IDS) should have higher retaining probability to propagation. However, IDS is a non-differentiable metrics and cannot be directly optimize. Inspired by the idea of learning to rank\cite{DBLP:conf/icml/BurgesSRLDHH05}, we could adopt the pair-wise ranking loss and encode the non-differentiable metrics in continuous functions.
\begin{equation}
\begin{aligned}
L^{rank}_{t} = \sum_{\hat{T}_i^t, \hat{T}_j^t \in \mathcal{T}_{t}} {\rm Sigmoid} \left( \gamma * \left(f_s\left(\hat{T}_i^t\right) - f_s\left(\hat{T}_j^t\right) \right)\right)\\
\begin{cases} 
\gamma = 1 \quad if \quad IDS\left(\hat{T}_i^t\right) > IDS\left(\hat{T}_i^t\right) \\
\gamma = -1 \quad if \quad IDS\left(\hat{T}_i^t\right) < IDS\left(\hat{T}_i^t\right) 
\end{cases}
\end{aligned}
\label{equation:rank_loss}
\end{equation}
 
where the $\gamma$ is the rank label of the pair-wise tracklets and the $IDS\left(\right)$ represent the IDS of a tracklet.  
Then, the total loss is,
\begin{equation}
L_{t} = L^{margin}_{t} + L^{rank}_{t}
\label{equation:loss}
\end{equation}
\textbf{Search-Based Tracklet Optimizing}
We now introduce an innovative algorithm for tracklet level training called Search-Based Tracklet Optimizing (SBTO). 
It avoids the aforementioned exposure bias problem as we iteratively involve the obfuscated candidate tracklets in training.

The overall architecture of our SBTO is illustrated in Figure \ref{fig.tracklet_search} and Algorithm \ref{alg:algorithm.tracklet_search}, which consists of five major steps: 

1. For a specific tracked object $o$, assuming we have $K$ retained tracklet proposals(after pruning) $\mathcal{T}_t = \{\hat{T}_{1}^{t},  \hat{T}_{2}^{t}, ..., \hat{T}_{K}^{t} \}$ at frame $t$, where $K$ is a constant we set to limit the number of retained tracklet proposals.

2. Applying candidate search to extend each tracklet proposals and build hypothesis propagation tree according to the detection results at frame $t+1$.  Here, let $C$ denotes the candidate number of each searched object. After this step, we get $K \times C$ hypothesis tracklet proposals in $t+1$,  $\mathcal{T}_{t+1}^{s} = \{\hat{T}_1^{t+1},  \hat{T}_2^{t+1}, ..., \hat{T}_{K \times C}^{t+1} \}$.

3. Calculating the score of each hypothesis tracklet proposals $\hat{T}_i^{t+1}$ and ground truth tracklet with the scoring function $f_s\left( T\right)$. In this paper, we use an encoder model to parameterize $f_s\left( T\right)$. Specifically, we implement encoder with sequence model to extract appearance feature of tracklet. (see details in section \ref{section:methods:online appearance encoder}). Afterwards, we rank the tracklet scores in a decreasing order. 

4. We use the ranked tracklet score to prune easy hypothesis tracklet proposals to constrain the number of proposals. Define the tracklet $\hat{T}_K^{t+1} \in \mathcal{T}_{t+1}^{s}$ to be the $K$'th ranked hypothesis tracklet proposal according the $f_s\left(T \right)$. We keep the top $K$ tracklet proposals: 
\begin{equation}
    \mathcal{T}_{t+1} = \left\{\hat{T}_i^{t+1} \in \mathcal{T}_{t+1}^{s} | f_s\left( \hat{T}_i^{t+1} \right) \geq f_s\left( \hat{T}_K^{t+1} \right) \right\}
\end{equation}

5. We now define the loss in frame $t+1$ as the summation of top $K$ hypothesis tracklet proposals' loss. Finally, to learn long-term dependency of tracklet in temporal, we accumulate losses in each step over time recurrently. Let $N$ denotes the total steps of tracklet, the total losses of each tracklet is:

\begin{equation}
L_{total} = \sum_{t=1}^{N-1} L_{t+1}
\end{equation}

\begin{figure}[t!]
\begin{center}
  \includegraphics[width=0.9\linewidth]{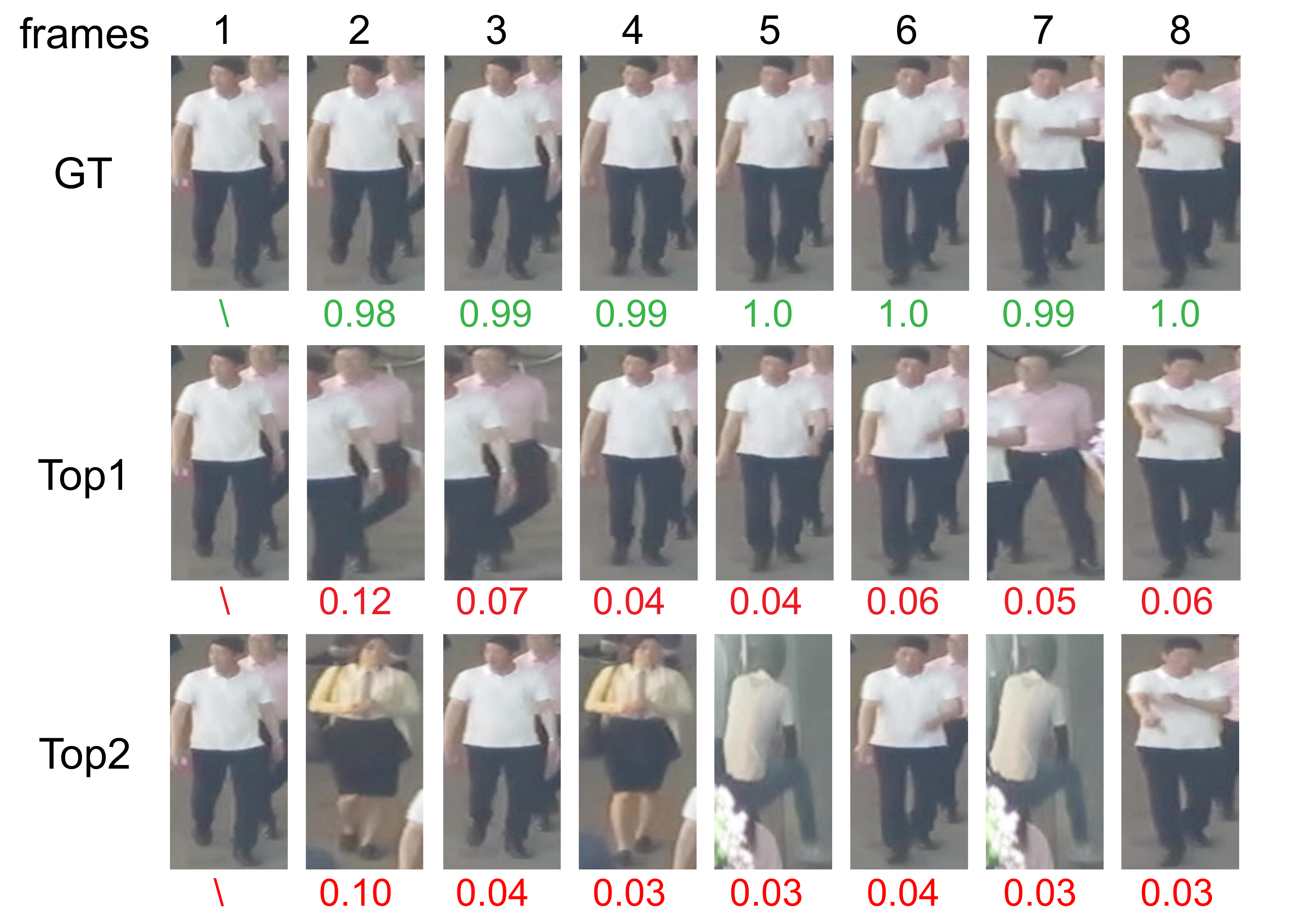}
\end{center}
\caption{The Top2 ranked hypothesis tracklet proposals in training, the number represents the tracklet
score at each time step.}
\label{fig.hard_mining}
\end{figure}

Unlike others standard network in training, SBTO requires running search to collect all Top$-K$ hypothesis tracklet proposals. In the forward process, we record $\mathcal{T}_{t}, L_{t}$ and the hidden states of sequence model that contribute to losses at each time step during propagating hypothesis tracklet. In the backward pass, we back-propagate the errors by adapting back-propagation through time (BPTT) algorithms~\cite{DBLP:journals/compsys/Mozer89}. 
As shown in Figure \ref{fig.hard_mining}, the online search could also mine the hard negative tracklets which are high similar with the ground truth tracklet, and assign it a low score after training through margin loss after several epochs in the training stage.

\subsection{Online Appearance Encoder}
\label{section:methods:online appearance encoder}
Visual tracking relies on temporal context, which motivates us to incorporate appearance feature from historical contents to enrich the tracklet representation. From this perspective, a direct method that could model temporal appearance information and long-term dependency is sequence model such as RNN. In our framework, we build the scoring function $f_s\left( T \right)$ by various sequence model to encode the appearance information of the tracklets, as shown in Fig. \ref{fig.overall.arch}.

For a tracked object $o$ with tracklet $T_i$, let bounding boxes $\{b_i^1, b_i^2, ..., b_i^t, b_i^{t+1}\}$ denote the location of $T_i$ at each frame $t$. We use Convolutional Neural Network (CNN) to extract the appearance of each bounding boxes. Especially, the CNN accepts the raw patch as input, and output the flattened last layer feature map of CNN. Let $\phi_i^1, \phi_i^2, ..., \phi_i^t,\phi_i^{t+1}$ denote appearance feature for patches at each time step, where $\phi_i^t$ is a vector with dimension $H$. Our online appearance encoder accepts the sequence of $\{\phi_i^1, \phi_i^2, ..., \phi_i^t,\phi_i^{t+1}\}$ as input and output hidden state vector $h_i^{t+1}$ with dimension $H$ recurrently.
\begin{equation}
h_i^{t+1} = f\left( \phi_i^{t+1}, h_i^{t} \right)
\end{equation}
The appearance encoder of tracklet could be built by sequence model. We have explored various sequence model to implement the encoder including LSTM~\cite{DBLP:journals/neco/HochreiterS97}, LSTM with attention~\cite{DBLP:journals/corr/BahdanauCB14} and transformer~\cite{DBLP:conf/nips/VaswaniSPUJGKP17}. In this section, We will use LSTM with attention as an example to illustrate the encoder. 
Define the context vector at time step $t$ as $c_i^{t+1}$, then it could be computed as a weighted sum of historical information:
\begin{equation}
c_i^{t+1} = \sum_{j=1}^{t+1} \alpha_{j} h_i^{j}
\end{equation}
Let ${h_i^{t+1}}^T$ represents the transpose of $h_i^{t+1}$, then $\alpha_{j}$ could be formulated as:
\begin{equation}
\alpha_{j} = \frac{\exp( h_i^{j} \cdot {h_i^{t+1}}^T )}{\sum_{j=1}^{t+1}{\exp( h_i^{j} \cdot {h_i^{t+1}}^T )}} 
\end{equation}
The output $c_i^{t+1}$ of attention layer at time step $t+1$ is then passed to another fully connected layer which condenses the $H$ dimensional vector to a scalar. 

In this section, we use sequence model to leverage all prior appearance information from a given tracklet. Note that compared with previous studies~\cite{DBLP:conf/iccv/KimLCR15,DBLP:conf/eccv/KimLR18,DBLP:conf/iccv/SadeghianAS17}, we only use the appearance information in our tracklet scoring network. 

\begin{algorithm}[ht] 
\caption{Search Based Tracklet Optimization.} 
\label{alg:algorithm.tracklet_search} 
\begin{algorithmic}[1]
\REQUIRE ~~\\ 
The number of retained tracklet proposal, $K$;\\
The candidate detections at each time step, $C_t$;\\
The tracker object, $o$;\\
The value of margin, $\alpha$;
\ENSURE ~~\\ 
Total loss, $L_{total}$;
\STATE $ \mathcal{T}_1 \leftarrow \{\hat{T}_{o,1}^{1},  \hat{T}_{o,2}^{1}, ..., \hat{T}_{o,K}^{1} \}$;
\STATE $L(T_{gt}, \hat{T}) \leftarrow max \left( 0, \alpha - f_s\left(T_{gt}\right) + f_s(\hat{T}) \right)$; 
\FOR{$\tau \leftarrow ~ 1 ~ to ~ t-1 $}
\STATE $ \mathcal{T}_{\tau+1} \leftarrow \left\{ \right\} $ ;\
\FOR{$T$ \textbf{in} $\mathcal{T}_{\tau}$}
\FOR{$c$ \textbf{in} $C_{\tau}$}
\STATE Append $c$ to $T$;\
\STATE Add $T$ to $\mathcal{T}_{\tau+1}$;\
\ENDFOR
\ENDFOR
\STATE $\mathcal{T}_{\tau+1} \leftarrow TopK \left(\{\mathcal{T}_{\tau+1} \backslash T_{gt}^{\tau+1}\}\right)$;\
\STATE $L_{\tau+1} \leftarrow \sum_{\hat{T_i}^{\tau+1} \in \{\mathcal{T}_{\tau+1} \backslash T_{gt}^{\tau+1}\}}{L(T_{gt}^{\tau+1}, \hat{T_i}^{\tau+1})} $;\
\ENDFOR
\STATE $L_{total} = \sum_{\tau=1}^{t-1} L_{\tau+1}$; 
\RETURN $L_{total}$; 
\end{algorithmic}
\end{algorithm}

\subsection{Application in Tracking}
\label{section:methods:application of SBTO to tracking}
Our method follows the online tracking-by-detection paradigm, which generates the trajectories by associating detection results across frames. To further validate the effeteness of SBTO for different association algorithms, we conduct data-association with Hungarian algorithm (Online) and multiple hypothesis tracking algorithm (MHT, near Online). Considered that the Hungarian algorithm is a general method in MOT. In this section, we briefly summarize the key steps related to multiple hypothesis tracking of our implementation. The key steps of MHT consist of hypothesis tree construction, gating, association with MWIS and pruning. More details about MHT can be found in~\cite{DBLP:conf/iccv/KimLCR15}.

\noindent
\textbf{Hypothesis Tree Construction.} 
For each target object, the hypothesis tree starts from the detection that it first appears, and
will be extended by appending children detection in the next frame. Each tree node in the hypothesis tree corresponds to one detection. 
Each path from root to leaf represents a candidate tracklet proposal. In this children spawning step, only detection within a gated region are considered. This process is repeated recurrently until the final hypothesis tree is constructed completely. During the tree construction, tracklet score based our proposed scoring function $f_s\left( T\right)$ of each hypothesis path are recorded and would be used for tree pruning later.

\noindent
\textbf{Gating and Association.}
To avoid combinatorial explosion in the spawn of tree generation, We need to gate out disturbing detections in next frames. We use IOU score between the $n$th detection $d_n^{t+1}$ with the last detection of tracklet proposal $d_T^{t1}$ as the criterion of gating. With detection chosen from gating, we can build up the hypothesis tree to run multiple hypothesis proposal propagation. Afterwards, we use MWIS to find the best set of tracks, with details refer to ~\cite{DBLP:conf/cvpr/BrendelAT11}.

\noindent
\textbf{Pruning} We follow the standard $N$-scan pruning approach to delete the conflicting hypothesis path. For each of selected path in frame $t$, we trace back to the node in frame $t-N$ and prune the sub-trees that is conflict with the selected path at that node. Note that a larger $N$ makes a large window to delay decision, which will bring an improvement in precision but take more time consumption. After pruning, only the surviving hypothesis path are updated in the next frame.

\section{Experiments}
\label{section:experiments}
In this section, we introduce the details of datasets, evaluation metric and the implementation details firstly, and then we perform an in-depth analysis of our method on various benchmark datasets on the MOT challenge. Finally, we present more insights and ablation studies on our proposed methods.

\subsection{Datasets and Metric}
\label{section:experiments:dasets and metric}
\noindent 
\textbf{Datasets} To test the power of our method, we report the quantitative results on the three datasets in MOT Challenge Benchmark~\cite{DBLP:journals/corr/Leal-TaixeMRRS15,DBLP:journals/corr/MilanL0RS16}. This benchmarks are widely used to evaluate the performance of multi-object trackers.

\textbf{2DMOT2015~\cite{DBLP:journals/corr/Leal-TaixeMRRS15}} It consists of 11 training sequences and 11 testing sequences. This dataset only contains 500 tracks in training sets, but is potentially more challenging because of the low resolution, quite noisy detections.

\textbf{MOT16~\cite{DBLP:journals/corr/MilanL0RS16}} It consists of 7 training sequences and 7 testing sequences with moving and stationary cameras in various pedestrians’ scenes. MOT16 provides the detection responses of DPM~\cite{DBLP:journals/computer/Forsyth14} for training and testing.

\textbf{MOT17} It contains the same videos as the MOT16 but with more precise annotation. 
Moreover, the sequences are provided with detection results from two more detection algorithms: Faster-RCNN~\cite{DBLP:journals/pami/RenHG017} and  SDP~\cite{DBLP:conf/cvpr/YangCL16}.

For a fair comparison, we use the public detection results provided with each datasets as the input of our approach.

\noindent 
\textbf{Evaluation Metric} For performance evaluation, we follow the standard CLEAR MOT metrics\cite{DBLP:conf/cvpr/MilanSR13a} used in MOT Benchmarks, which consist of Multiple Object Tracking Accuracy (MOTA), Multiple Object Tracking Precision (MOTP) Mostly Track targets (MT), Mostly Lost targets (ML), False Positives (FP), False Negatives (FN), ID Switches (IDS), ID F1 Score (IDF1), ID Precision (IDP), ID Recall (IDR), Fragment errors(Frag). Detailed descriptions about these metrics can be found in \cite{DBLP:conf/cvpr/MilanSR13a}. 

\subsection{Implementation Details}
\label{section:experiments:implementation details}
\noindent 
\textbf{Network Architecture} We use pre-trained ResNet-50~\cite{DBLP:conf/cvpr/HeZRS16} with ImageNet Image Classification task~\cite{DBLP:journals/ijcv/RussakovskyDSKS15} as our backbone network and then finetune this model on the MOT training datasets. The output feature map of the last convolution layer of ResNet-50 is fed in an embedding network. The embedding network consists of a convolution layer with 256 output channels and a fully connected layer with output dimension 256, which are used to reduce the channel dimension and generate the final appearance feature, respectively.

Given the bounding boxes of detection, we crop and resize it to the size of $128 \times 64$, and feed it into the backbone and embedding network to generate appearance feature. The tracklet encoder is built by LSTM with attention. A single layer LSTM with hidden size 256 is used to model the temporal information and construct tracklet. The final step hidden state of LSTM used to calculate the context feature with the previous hidden state through attention mechanism. Then the context feature is fed to a single-layer fully connected network to generate the tracklet score.

\noindent 
\textbf{Tracklet Proposals} Due to the memory limit of GPUs, we construct artificial tracklet proposals with a maximum length of $N_{length}$  as the training data. First, we randomly pick one ground truth tracklet of length $N_{length}$ from the annotations, which is a clip of the whole trajectories. For every frame in this tracklet proposal, we randomly sample $N_{candidates}$ other bounding boxes as the candidates in the process of tracklet hypothesis generation. Finally, we construct $N_{length} \times N_{candidates}$ patches as a batch of input for each training iteration.

\noindent 
\textbf{Training} During training, we apply the Adam optimizer~\cite{DBLP:journals/corr/KingmaB14} to end-to-end train the network and set the weight decay rate to $5e-4$. We used 5 epochs for model warming up and then train another 45 epochs at a learning rate of $1e^{-5}$. The margin $\alpha$ is set to 1. The batch size is set to 16. The maximum length $N_{max}$ and the candidate number $N_{candidates}$ of the best model is 8 and 8, respectively.

\noindent 
\textbf{Inference} During inference, following the common practice of online tracking approaches~\cite{DBLP:journals/corr/abs-1808-01562}, we pre-processing the original detection results. 
The scores of generated candidate tracklets in each frame step are computed according to section \ref{section:methods:tracklet level optimization}. The association is then achieved by solving the bipartite graph (online) or MHT (near on-line). We set the hyper-parameters pruning $K$ as 3 in MHT. 

\noindent 
\textbf{Platform} All experiments are conducted on a 1.2GHZ Intel Xeon server with 8 NVIDIA TITAN X GPUs. The deep learning framework we used is Pytorch.

\subsection{Comparison to the State of the Art}
\label{section:experiments:comparison to the state of the art}
We use public detection results for comparison. To further validate the effectiveness of our methods for different association algorithms, we conduct experiments with the Hungarian algorithm (Online) and MHT (Near Online). We report the performance comparison with other SOTA methods on the MOT Challenge 15/16/17 Benchmark in Table \ref{tab.2DMOT2015},\ref{tab.MOT2016},\ref{tab.MOT2017} respectively. For a fair comparison, we select most recently  published trackers such as STRN\cite{DBLP:journals/corr/abs-1904-11489}, FAMNet\cite{DBLP:journals/corr/abs-1904-04989} among online trackers, and almost all MHT-based trackers. Besides, we add some offline trackers for comprehensiveness, such as SAS\cite{DBLP:conf/cvpr/MaksaiF19} which also tries to solve the problem of exposure bias. Note that our method only uses appearance feature in the scoring model. To the best of our knowledge, our tracker achieves the most promising result with similar setting. In online setting, our method achieve MOTA 40.0, 50.1, 52.6 on MOT15, MOT16 and MOT17, respectively, which beats almost all online methods in recently published results. And we have obtained better results in our near-online version tracker based on MHT, which outperforms all other MHT based methods in three benchmark datasets with the major metric MOTA and IDF. Compared to \cite{DBLP:conf/cvpr/MaksaiF19}, which tends to drop many short expected trajectories and leads to higher FN, our neutral measurement of tracklet quality can balance various MOT metrics without sacrificing MOTA.

\begin{table*}[ht!]
\small
\begin{center}
\begin{tabular}{c|c|c c c c c c c c c c c}
\hline
Mode & Method & MOTA$\uparrow$ & MOTP$\uparrow$ & FP$\downarrow$ & FN$\downarrow$ & IDF$\uparrow$ & IDP$\uparrow$ & IDR$\uparrow$ & IDS$\downarrow$ & Frag$\downarrow$ & MT$\uparrow$ & ML$\downarrow$ \\
\hline
Offline & JointMC\cite{keuper2018motion} & 35.6 & 71.9 & 10,580 & 28,508 & 45.1 & 54.4 & 38.5 & 457 & 969 & 23.2 & 39.3\\
Offline & SAS(motion only)\cite{DBLP:conf/cvpr/MaksaiF19} & 22.2 & 71.1 & 5,591 & 41,531 & 27.2 & 46.3 & 19.2 & 700 & 1,240 & 3.1 & 61.6 \\
\hline
Online & AMIR~\cite{DBLP:conf/iccv/SadeghianAS17} & 37.6 & 71.7 & 7,933 & 29,397 & 46.0 & 58.4 & 38.0 & \textcolor[rgb]{1,0,0}{1,026} & 2,024 & 15.8 & 26.8\\
Online & STRN~\cite{DBLP:journals/corr/abs-1904-11489} & 38.1 & 72.1 & \textcolor[rgb]{1,0,0}{5,451} & 31,571 & \textcolor[rgb]{1,0,0}{46.6} & \textcolor[rgb]{1,0,0}{63.9} & 36.7 & 1,033 & 2,655 & \textcolor[rgb]{1,0,0}{33.4} & \textcolor[rgb]{1,0,0}{11.5} \\
Online & \textbf{Ours} & \textcolor[rgb]{1,0,0}{40.0} & \textcolor[rgb]{1,0,0}{73.4} & 9,349 & \textcolor[rgb]{1,0,0}{26,328} & 44.3 & 52.7 & \textcolor[rgb]{1,0,0}{38.1} & 1,207 & \textcolor[rgb]{1,0,0}{1,624} & 17.1 & 28.8\\
\hline
Near online & MHT\_DAM~\cite{DBLP:conf/iccv/KimLCR15} & 32.4 & 71.8 & 9,064 & 32,060 & 45.3 & \textcolor[rgb]{1,0,0}{58.9} & 36.8 & \textcolor[rgb]{1,0,0}{435} & \textcolor[rgb]{1,0,0}{826} & \textcolor[rgb]{1,0,0}{16.0} & 43.8\\
Near online & \textbf{Ours} & \textcolor[rgb]{1,0,0}{41.3} &\textcolor[rgb]{1,0,0}{73.5} & \textcolor[rgb]{1,0,0}{8,000} & \textcolor[rgb]{1,0,0}{27,210} & \textcolor[rgb]{1,0,0}{46.1} & 56.6 & \textcolor[rgb]{1,0,0}{38.9} & 852 & 1,405 & 15.7 & \textcolor[rgb]{1,0,0}{34.5}\\
\hline
\end{tabular}
\end{center}
\caption{Tracking Performance on 2DMOT2015 benchmark dataset.}
\label{tab.2DMOT2015}
\end{table*}

\begin{table*}[ht!]
\small
\begin{center}
\begin{tabular}{c|c|c c c c c c c c c c c}
\hline
Mode & Method & MOTA$\uparrow$ & MOTP$\uparrow$ & FP$\downarrow$ & FN$\downarrow$ & IDF$\uparrow$ & IDP$\uparrow$ & IDR$\uparrow$ & IDS$\downarrow$ & Frag$\downarrow$ & MT$\uparrow$ & ML$\downarrow$ \\
\hline
Offline & LMP\cite{DBLP:conf/cvpr/TangAAS17} & 48.8 & 79.0 & 6,654 & 86,245 & 51.3 & 71.1 & 40.1 & 481 & 595 & 18.2 & 40.1\\
Offline & FWT\cite{DBLP:journals/corr/HenschelLCR17} & 47.8 & 75.5 & 8,886 & 85,487 & 44.3 & 60.3 & 35 & 852 & 1,534 & 19.1 & 38.2\\
\hline
Online & MOTDT~\cite{DBLP:conf/icmcs/ChenAZS18} & 47.6 & 74.8 & 9,253 & 85,431 & \textcolor[rgb]{1,0,0}{50.9} & \textcolor[rgb]{1,0,0}{69.2} & \textcolor[rgb]{1,0,0}{40.3} & 792 & 1,858 & 15.2 & \textcolor[rgb]{1,0,0}{38.3}\\
Online & AMIR~\cite{DBLP:conf/iccv/SadeghianAS17} & 47.2 & 75.8 & \textcolor[rgb]{1,0,0}{2,681} & 92,856 & 46.3 & 68.9 & 34.8 & \textcolor[rgb]{1,0,0}{774} & 1,675 & 14.0 & 41.6\\
Online & \textbf{Ours} & \textcolor[rgb]{1,0,0}{50.1} & \textcolor[rgb]{1,0,0}{76.5} & 5,582 & \textcolor[rgb]{1,0,0}{84,629} & 48.1 & 66.5 & 37.6 & 786 & \textcolor[rgb]{1,0,0}{1,294} & \textcolor[rgb]{1,0,0}{16.3} & 40.7\\
\hline
Near online & MHT\_bLSTM~\cite{DBLP:conf/eccv/KimLR18} & 42.1 & 75.9 & 11,637 & 93,172 & 47.8 & \textcolor[rgb]{1,0,0}{67.2} & 37.1 & 753 & 1,156 & 14.9 & 44.4 \\
Near online & EDMT~\cite{DBLP:conf/cvpr/ChenSZX17} & 45.3 & 75.9 & 11,122 & 87,890 & 47.9 & 65.3 & 37.8 & 639 & 946 & 17.0 & \textcolor[rgb]{1,0,0}{39.9}\\
Near online & MHT\_DAM~\cite{DBLP:conf/iccv/KimLCR15} & 45.8 & \textcolor[rgb]{1,0,0}{76.3} & \textcolor[rgb]{1,0,0}{6,412} & 91,758 & 46.1 & 66.3 & 35.3 & \textcolor[rgb]{1,0,0}{590} & \textcolor[rgb]{1,0,0}{781} & 16.2 & 43.2\\
Near online & \textbf{Ours} & \textcolor[rgb]{1,0,0}{50.4} &\textcolor[rgb]{1,0,0}{76.3} & 8,491 & \textcolor[rgb]{1,0,0}{81,156} & \textcolor[rgb]{1,0,0}{50.1} & 66.7 & \textcolor[rgb]{1,0,0}{40.1} & 807 & 1,251 & \textcolor[rgb]{1,0,0}{17.4} & \textcolor[rgb]{1,0,0}{39.9}\\
\hline
\end{tabular}
\end{center}
\caption{Tracking Performance on MOT2016 benchmark dataset.}
\label{tab.MOT2016}
\end{table*}

\begin{table*}[ht!]
\small
\begin{center}
\begin{tabular}{c|c|c c c c c c c c c c c}
\hline
Mode & Method & MOTA$\uparrow$ & MOTP$\uparrow$ & FP$\downarrow$ & FN$\downarrow$ & IDF$\uparrow$ & IDP$\uparrow$ & IDR$\uparrow$ & IDS$\downarrow$ & Frag$\downarrow$ & MT$\uparrow$ & ML$\downarrow$ \\
\hline
Offline & FWT\cite{DBLP:journals/corr/HenschelLCR17} & 51.3 & 77 & 24,101 & 247,921 & 47.6 & 63.2 & 38.1 & 2,648 & 4,279 & 21.4 & 35.2\\
Offline & SAS\cite{DBLP:conf/cvpr/MaksaiF19} & 44.2 & 76.4 & 29,473 & 283,611 & 57.2 & 80.6 & 44.3 & 1,529 & 2,644 & 16.1 & 44.3 \\
\hline
Online & STRN~\cite{DBLP:journals/corr/abs-1904-11489} & 50.9 & 75.6 & 27,532 & 246,924 & \textcolor[rgb]{1,0,0}{56.5} & \textcolor[rgb]{1,0,0}{74.5} & \textcolor[rgb]{1,0,0}{45.5} & 2,593 & 9,622 & \textcolor[rgb]{1,0,0}{20.1} & 37.0\\
Online & FAMNet\cite{DBLP:journals/corr/abs-1904-04989} & 52.0 & 76.5 & \textcolor[rgb]{1,0,0}{14,138} & 253,616 & 48.7 & 66.7 & 38.4 & 3,072 & 5,318 & 19.1 & \textcolor[rgb]{1,0,0}{33.4} \\
Online & \textbf{Ours} & \textcolor[rgb]{1,0,0}{52.6} & \textcolor[rgb]{1,0,0}{76.6} & 20,089 & \textcolor[rgb]{1,0,0}{244,930} & 51.3 & 68.3 & 41.1 & \textcolor[rgb]{1,0,0}{2,530} & \textcolor[rgb]{1,0,0}{4,170} & 19.5 & 38.2 \\
\hline
Near online & EDMT~\cite{DBLP:conf/cvpr/ChenSZX17} & 50.0 & 77.3 & 32,279 & 247,297 & 51.3 & 67.0 & 41.5 & \textcolor[rgb]{1,0,0}{2,264} & 3,260 & \textcolor[rgb]{1,0,0}{21.6} & \textcolor[rgb]{1,0,0}{36.3} \\
Near online & MHT\_DAM~\cite{DBLP:conf/iccv/KimLCR15} & 50.7 & \textcolor[rgb]{1,0,0}{77.5} & 22,875 & 252,889 & 47.2 & 63.4 & 37.6 & 2,314 & \textcolor[rgb]{1,0,0}{2,865} & 20.8 & 36.9\\
Near online & \textbf{Ours} & \textcolor[rgb]{1,0,0}{53.3} & 76.5 & 
\textcolor[rgb]{1,0,0}{22,161} & \textcolor[rgb]{1,0,0}{238,959} & \textcolor[rgb]{1,0,0}{53.3} & \textcolor[rgb]{1,0,0}{70.0} & \textcolor[rgb]{1,0,0}{43.1} & 2,434 & 4,089 & 20.0 & 38.7\\
\hline
\end{tabular}
\end{center}
\caption{Tracking Performance on MOT2017 benchmark dataset.}
\label{tab.MOT2017}
\end{table*}

\subsection{Ablation Study}
\label{section:experiments:ablation study}
We now present a transparent demonstration of the impact of each component we have introduced towards these two goals. We perform comparative experiments on MOT17 and report the tracking results of the minimal validation loss model. For all experiments in this section, we split three sequences from training sets ((MOT17-02, MOT17-05, MOT17-09 of DPM, FRCNN, and SDP three detectors) for validation and the rest for training.  And we report the results of our online association method (by Hungarian algorithm) for better illustration.

\noindent 
\textbf{Impact of the margin loss and online tracklet search.} We first investigate the contribution of each components in our methods by measuring the performance on the validation set. We conduct baseline experiments on three variants of our model.
In the EXP1, we replace the margin loss and rank loss (Equation \ref{equation:loss}) with cross-entropy loss as our baseline. And remove the online hypothesis tracklet search in the training phase, i.e., the positive/negative tracklets are randomly sampled in training sets. To avoid the LSTM overfitting to a fixed-length sequence, we construct variable-length artificial track proposals which are generated from ground truth track annotations as in \cite{DBLP:conf/eccv/KimLR18}. In the EXP2, we keep the margin loss but remove rank loss and tracklet search as EXP1. In the EXP3, we only remove the tracklet search in our original model. For a fair comparison, we set the maximum sequence length $N_{length}$  to 8 in three experiments. As shown in Table  \ref{tab.ablation_study_baseline}, with margin loss and rank loss, the tracking accuracy improves by 6.8 and 2.1 in terms of MOTA compared to control respectively. It is worth to point out our proposed margin loss and rank loss reduce FP and IDS significantly, which indicates that this cost function could help tracker to identify incorrect association more accurately. Compared to EXP3, our methods achieve an additional 2.5 MOTA and -259 IDS improvement, which is in line with our expectations that the online tracklet search could reduce exposure bias.

\begin{table*}[h]
\small
\begin{center}
\begin{tabular}{c  c | c | c |c c c c c c}
\hline
 & Margin loss & Rank loss & Tracklet search & MOTA$\uparrow$ & FP$\downarrow$ & FN$\downarrow$ & IDF$\uparrow$ & Frag$\downarrow$ & IDS$\downarrow$ \\
\hline
EXP1 & $\times$ & $\times$ & $\times$ & 34.8 & 8,600 & 50,888 & 36.1 & 1,083 & 797 \\
EXP2 & \checkmark & $\times$ & $\times$ & 41.6 & 1,817 & 51,413 & 43.2 & 936 & 756 \\
EXP3 & \checkmark & \checkmark & $\times$ & 43.7 & 1632 & 50,033 & 44.8 & 758 & 633 \\
Ours & \checkmark & \checkmark & \checkmark & \textbf{46.2} & \textbf{1,141} & \textbf{48,196} & \textbf{47.3} & \textbf{512}  & \textbf{374} \\
\hline
\end{tabular}
\end{center}
\caption{The effect of each module in our design. The loss function design and tracklet search bring a large gain in tracking performance.}
\label{tab.ablation_study_baseline}
\end{table*}

\begin{figure}[ht!]
\small
\begin{center}
  \includegraphics[width=1.0\linewidth]{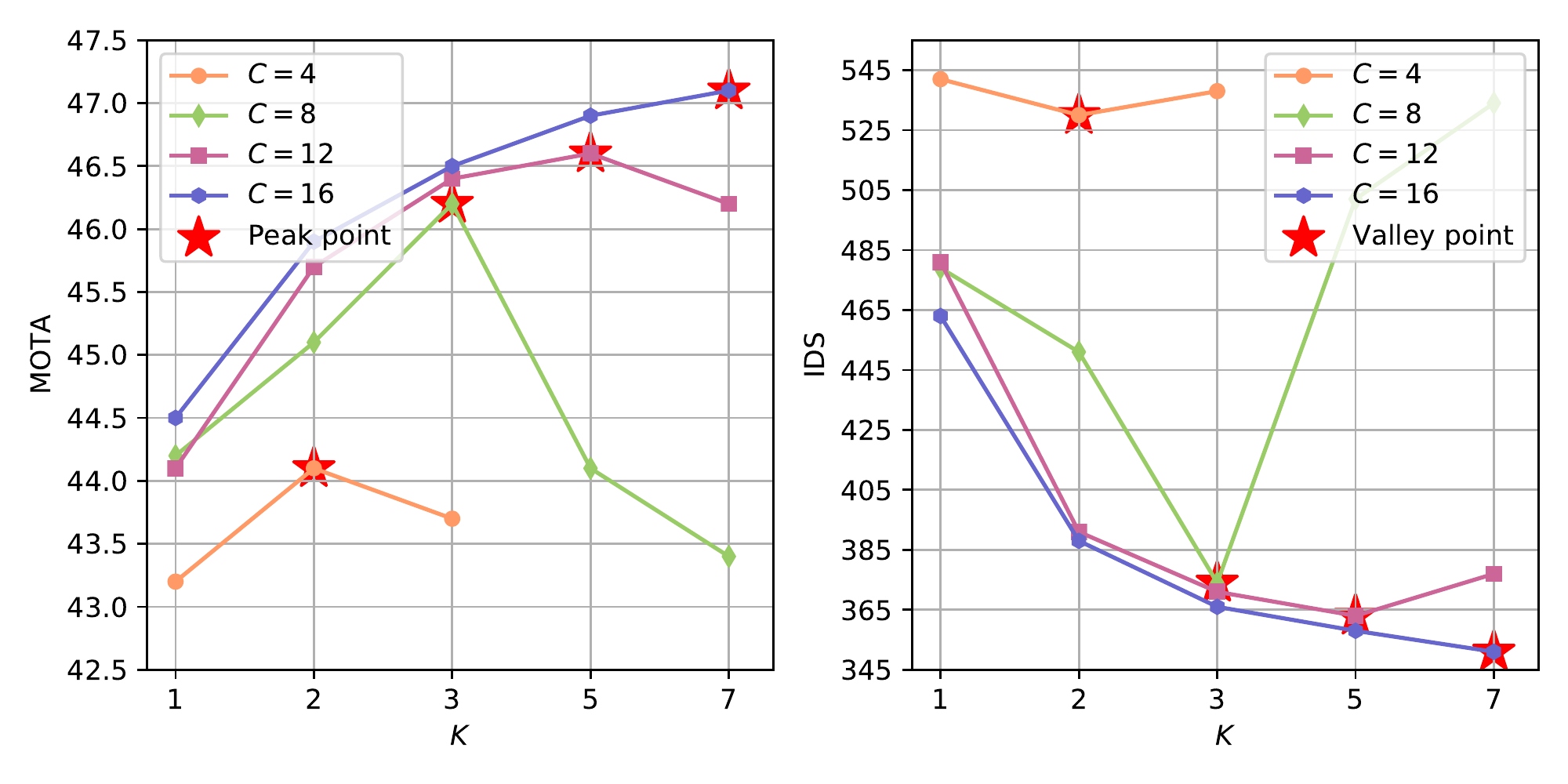}
\end{center}
\caption{Increasing $K$ and $C$ simultaneously can improve MOTA and reduce IDS.}
\label{fig.ablation_study_parameters}
\end{figure}

\noindent 
\textbf{Impact of hyper-parameters in online tracklet search.} Then, we perform a sensitivity analysis and examine the effects of the various configuration of hyper-parameters in cost calculation and online hypothesis tracklet search. As we explained earlier, The number of retained tracklet proposal $K$ is a central parameter in our methods. It is preferable to use large $K$ so that sufficient and diversified tracklet could be sampled. In contrast, a large $K$ will introduce easy tracklet for a given limited $C$, which could lead to model convergence prematurely. Another pivotal parameter is the candidate number $C$. We tend to use large $C$ to search indistinguishable detections and generate hypothesis tracklet which could not be pruned early. However, we cannot increase $C$ unrestrictedly with the limit of GPU memory. Figure \ref{fig.ablation_study_parameters} shows the results from this analysis over different parameters of $K$ and $C$. In these experiments, we set the maximum sequence length $N_{length}$  to 8, and all other parameters are consistent in training and inference stages. The results show that $C$ is positively affecting the tracking accuracy. It is intuitive since large $C$ could introduce sufficient training sample as we analyzed above. On the other hand, the too small or too big $K$ have resulted in a decrease in tracking accuracy when $C$ is fixed, which is also accord with our hypothesis. To summarize, Figure \ref{fig.ablation_study_parameters} indicates that expanding the search space by increasing both $ C $ and $ K $ simultaneously will improve tracking performance.

\noindent 
\textbf{Effectiveness of sequence model.}
As described above, the LSTM with attention is just one implementation of the tracklet encoder. In this section, we have explored the impact of different types of sequence models on tracking performance. To eliminate the influence of the sequence length of training, we set the sequence length equal to 8 in training/inference for different model and compare the tracking performance on the validation sets. As shown in Table \ref{table.effect_sequence_model}, all sequence model, including LSTM, transformer and LSTM with attention have achieved performance boosting  compared to the baseline. This results suggest that our learning framework can reduce exposure bias effectively, and the tracklet encoder is pluggable and can be extended to more powerful models. Our results also take the leading in finding the advantages of using LSTM with attention, instead  of  LSTM,  which  is  the  common practice in MOT area~\cite{DBLP:conf/nips/BengioVJS15,DBLP:conf/eccv/KimLR18,DBLP:conf/iccv/SadeghianAS17}. 

\begin{table*}[h]
\begin{center}
\begin{tabular}{l c c c |c|c|c}
\hline
& Margin loss & Rank loss & Tracklet search & LSTM  & Transformer & LSTM + attention  \\
\hline
EXP1 & $\times$ & $\times$ & $\times$ & 33.9 (828) & 35.3 (788) & 34.8 (797) \\
EXP2 & \checkmark & $\times$ & $\times$ & 40.2 (793) & 41.3 (762) & 41.6 (756) \\
EXP3 & \checkmark & \checkmark & $\times$ & 42.6 (667) & 42.9 (656) & 43.7 (633) \\
Ours & \checkmark & \checkmark & \checkmark & 45.1 (415) & 45.7 (392) & 46.2 (374) \\
\hline
\end{tabular}
\end{center}
\caption{The effect of different tracklet encoders on performance compared to baseline. The number in table represents MOTA$\uparrow$ (IDS$\downarrow$)}

\label{table.effect_sequence_model}
\end{table*}

\section{Conclusion}
\label{section:conclusion}
As a fundamental problem in sequence generation and association, exposure bias has attached the attention of many researchers. However, exposure bias is still an open problem in the MOT community, unlike other tasks in natural language processing. In this paper, we propose a novel method for optimizing tracklet consistency that directly takes into account the prediction errors in the training stage, which could eliminate exposure bias effectively. Second, our methods directly optimize the whole tracklet score but not frame-level cost, which is a more suitable model than pairwise matching. Experiments demonstrate that our approach can improve overall performance and achieve SOTA results in MOT challenge benchmarks. Our methods bring MOT an extra step closer to solve the training-inference mismatch problem. In future work, we will extend this approach to much larger application scenarios, like vehicle tracking, traffic light tracking, multi-camera multi-object tracking, which could further unlock the potential of our method.

{\small
\bibliographystyle{ieee_fullname}
\bibliography{egpaper_final}

\begin{thebibliography}{10}\itemsep=-1pt

\bibitem{DBLP:journals/corr/BahdanauCB14}
Dzmitry Bahdanau, Kyunghyun Cho, and Yoshua Bengio.
\newblock Neural machine translation by jointly learning to align and
  translate.
\newblock In {\em 3rd International Conference on Learning Representations,
  {ICLR} 2015, San Diego, CA, USA, May 7-9, 2015, Conference Track
  Proceedings}, 2015.

\bibitem{DBLP:conf/nips/BengioVJS15}
Samy Bengio, Oriol Vinyals, Navdeep Jaitly, and Noam Shazeer.
\newblock Scheduled sampling for sequence prediction with recurrent neural
  networks.
\newblock In {\em Advances in Neural Information Processing Systems 28: Annual
  Conference on Neural Information Processing Systems 2015, December 7-12,
  2015, Montreal, Quebec, Canada}, pages 1171--1179, 2015.

\bibitem{DBLP:conf/icip/BewleyGORU16}
Alex Bewley, ZongYuan Ge, Lionel Ott, Fabio~Tozeto Ramos, and Ben Upcroft.
\newblock Simple online and realtime tracking.
\newblock In {\em 2016 {IEEE} International Conference on Image Processing,
  {ICIP} 2016, Phoenix, AZ, USA, September 25-28, 2016}, pages 3464--3468,
  2016.

\bibitem{DBLP:conf/avss/BochinskiES17}
Erik Bochinski, Volker Eiselein, and Thomas Sikora.
\newblock High-speed tracking-by-detection without using image information.
\newblock In {\em 14th {IEEE} International Conference on Advanced Video and
  Signal Based Surveillance, {AVSS} 2017, Lecce, Italy, August 29 - September
  1, 2017}, pages 1--6, 2017.

\bibitem{DBLP:conf/cvpr/BrendelAT11}
William Brendel, Mohamed~R. Amer, and Sinisa Todorovic.
\newblock Multiobject tracking as maximum weight independent set.
\newblock In {\em The 24th {IEEE} Conference on Computer Vision and Pattern
  Recognition, {CVPR} 2011, Colorado Springs, CO, USA, 20-25 June 2011}, pages
  1273--1280, 2011.

\bibitem{DBLP:conf/icml/BurgesSRLDHH05}
Christopher J.~C. Burges, Tal Shaked, Erin Renshaw, Ari Lazier, Matt Deeds,
  Nicole Hamilton, and Gregory~N. Hullender.
\newblock Learning to rank using gradient descent.
\newblock In {\em Machine Learning, Proceedings of the Twenty-Second
  International Conference {(ICML} 2005), Bonn, Germany, August 7-11, 2005},
  pages 89--96, 2005.

\bibitem{DBLP:conf/cvpr/ChenSZX17}
Jiahui Chen, Hao Sheng, Yang Zhang, and Zhang Xiong.
\newblock Enhancing detection model for multiple hypothesis tracking.
\newblock In {\em 2017 {IEEE} Conference on Computer Vision and Pattern
  Recognition Workshops, {CVPR} Workshops 2017, Honolulu, HI, USA, July 21-26,
  2017}, pages 2143--2152, 2017.

\bibitem{DBLP:phd/dnb/Chen15a}
Lili Chen.
\newblock {\em Multiple People Tracking-by-Detection in a Multi-camera
  Environment}.
\newblock PhD thesis, Technical University Munich, 2015.

\bibitem{DBLP:conf/icmcs/ChenAZS18}
Long Chen, Haizhou Ai, Zijie Zhuang, and Chong Shang.
\newblock Real-time multiple people tracking with deeply learned candidate
  selection and person re-identification.
\newblock In {\em 2018 {IEEE} International Conference on Multimedia and Expo,
  {ICME} 2018, San Diego, CA, USA, July 23-27, 2018}, pages 1--6, 2018.

\bibitem{DBLP:journals/corr/abs-1904-04989}
Peng Chu and Haibin Ling.
\newblock Famnet: Joint learning of feature, affinity and multi-dimensional
  assignment for online multiple object tracking.
\newblock {\em CoRR}, abs/1904.04989, 2019.

\bibitem{DBLP:conf/iccv/ChuOLWLY17}
Qi Chu, Wanli Ouyang, Hongsheng Li, Xiaogang Wang, Bin Liu, and Nenghai Yu.
\newblock Online multi-object tracking using cnn-based single object tracker
  with spatial-temporal attention mechanism.
\newblock In {\em {IEEE} International Conference on Computer Vision, {ICCV}
  2017, Venice, Italy, October 22-29, 2017}, pages 4846--4855, 2017.

\bibitem{DBLP:journals/corr/abs-1907-12740}
Gioele Ciaparrone, Francisco~Luque S{\'{a}}nchez, Siham Tabik, Luigi Troiano,
  Roberto Tagliaferri, and Francisco Herrera.
\newblock Deep learning in video multi-object tracking: {A} survey.
\newblock {\em CoRR}, abs/1907.12740, 2019.

\bibitem{DBLP:journals/computer/Forsyth14}
David~A. Forsyth.
\newblock Object detection with discriminatively trained part-based models.
\newblock {\em {IEEE} Computer}, 47(2):6--7, 2014.

\bibitem{DBLP:journals/corr/Goodfellow17}
Ian~J. Goodfellow.
\newblock {NIPS} 2016 tutorial: Generative adversarial networks.
\newblock {\em CoRR}, abs/1701.00160, 2017.

\bibitem{DBLP:conf/nips/HeLXQ0L17}
Di He, Hanqing Lu, Yingce Xia, Tao Qin, Liwei Wang, and Tie{-}Yan Liu.
\newblock Decoding with value networks for neural machine translation.
\newblock In {\em Advances in Neural Information Processing Systems 30: Annual
  Conference on Neural Information Processing Systems 2017, 4-9 December 2017,
  Long Beach, CA, {USA}}, pages 178--187, 2017.

\bibitem{DBLP:conf/cvpr/HeZRS16}
Kaiming He, Xiangyu Zhang, Shaoqing Ren, and Jian Sun.
\newblock Deep residual learning for image recognition.
\newblock In {\em 2016 {IEEE} Conference on Computer Vision and Pattern
  Recognition, {CVPR} 2016, Las Vegas, NV, USA, June 27-30, 2016}, pages
  770--778, 2016.

\bibitem{DBLP:journals/corr/HenschelLCR17}
Roberto Henschel, Laura Leal{-}Taix{\'{e}}, Daniel Cremers, and Bodo Rosenhahn.
\newblock Improvements to frank-wolfe optimization for multi-detector
  multi-object tracking.
\newblock {\em CoRR}, abs/1705.08314, 2017.

\bibitem{DBLP:journals/neco/HochreiterS97}
Sepp Hochreiter and J{\"{u}}rgen Schmidhuber.
\newblock Long short-term memory.
\newblock {\em Neural Computation}, 9(8):1735--1780, 1997.

\bibitem{DBLP:journals/corr/HuangYDY15}
Lichao Huang, Yi Yang, Yafeng Deng, and Yinan Yu.
\newblock Densebox: Unifying landmark localization with end to end object
  detection.
\newblock {\em CoRR}, abs/1509.04874, 2015.

\bibitem{keuper2018motion}
Margret Keuper, Siyu Tang, Bjorn Andres, Thomas Brox, and Bernt Schiele.
\newblock Motion segmentation \& multiple object tracking by correlation
  co-clustering.
\newblock {\em IEEE transactions on pattern analysis and machine intelligence},
  2018.

\bibitem{DBLP:conf/iccv/KimLCR15}
Chanho Kim, Fuxin Li, Arridhana Ciptadi, and James~M. Rehg.
\newblock Multiple hypothesis tracking revisited.
\newblock In {\em 2015 {IEEE} International Conference on Computer Vision,
  {ICCV} 2015, Santiago, Chile, December 7-13, 2015}, pages 4696--4704, 2015.

\bibitem{DBLP:conf/eccv/KimLR18}
Chanho Kim, Fuxin Li, and James~M. Rehg.
\newblock Multi-object tracking with neural gating using bilinear {LSTM}.
\newblock In {\em Computer Vision - {ECCV} 2018 - 15th European Conference,
  Munich, Germany, September 8-14, 2018, Proceedings, Part {VIII}}, pages
  208--224, 2018.

\bibitem{DBLP:journals/corr/KingmaB14}
Diederik~P. Kingma and Jimmy Ba.
\newblock Adam: {A} method for stochastic optimization.
\newblock In {\em 3rd International Conference on Learning Representations,
  {ICLR} 2015, San Diego, CA, USA, May 7-9, 2015, Conference Track
  Proceedings}, 2015.

\bibitem{DBLP:journals/corr/Leal-TaixeMRRS15}
Laura Leal{-}Taix{\'{e}}, Anton Milan, Ian~D. Reid, Stefan Roth, and Konrad
  Schindler.
\newblock Motchallenge 2015: Towards a benchmark for multi-target tracking.
\newblock {\em CoRR}, abs/1504.01942, 2015.

\bibitem{DBLP:conf/eccv/LiZG18a}
Minxian Li, Xiatian Zhu, and Shaogang Gong.
\newblock Unsupervised person re-identification by deep learning tracklet
  association.
\newblock In {\em Computer Vision - {ECCV} 2018 - 15th European Conference,
  Munich, Germany, September 8-14, 2018, Proceedings, Part {IV}}, pages
  772--788, 2018.

\bibitem{DBLP:conf/icmcs/MaYYZZJX18}
Cong Ma, Changshui Yang, Fan Yang, Yueqing Zhuang, Ziwei Zhang, Huizhu Jia, and
  Xiaodong Xie.
\newblock Trajectory factory: Tracklet cleaving and re-connection by deep
  siamese bi-gru for multiple object tracking.
\newblock In {\em 2018 {IEEE} International Conference on Multimedia and Expo,
  {ICME} 2018, San Diego, CA, USA, July 23-27, 2018}, pages 1--6, 2018.

\bibitem{DBLP:conf/cvpr/MaksaiF19}
Andrii Maksai and Pascal Fua.
\newblock Eliminating exposure bias and metric mismatch in multiple object
  tracking.
\newblock In {\em {IEEE} Conference on Computer Vision and Pattern Recognition,
  {CVPR} 2019, Long Beach, CA, USA, June 16-20, 2019}, pages 4639--4648, 2019.

\bibitem{DBLP:journals/corr/MilanL0RS16}
Anton Milan, Laura Leal{-}Taix{\'{e}}, Ian~D. Reid, Stefan Roth, and Konrad
  Schindler.
\newblock {MOT16:} {A} benchmark for multi-object tracking.
\newblock {\em CoRR}, abs/1603.00831, 2016.

\bibitem{DBLP:conf/aaai/MilanRD0S17}
Anton Milan, Seyed~Hamid Rezatofighi, Anthony~R. Dick, Ian~D. Reid, and Konrad
  Schindler.
\newblock Online multi-target tracking using recurrent neural networks.
\newblock In {\em Proceedings of the Thirty-First {AAAI} Conference on
  Artificial Intelligence, February 4-9, 2017, San Francisco, California,
  {USA}}, pages 4225--4232, 2017.

\bibitem{DBLP:conf/cvpr/MilanSR13a}
Anton Milan, Konrad Schindler, and Stefan Roth.
\newblock Challenges of ground truth evaluation of multi-target tracking.
\newblock In {\em {IEEE} Conference on Computer Vision and Pattern Recognition,
  {CVPR} Workshops 2013, Portland, OR, USA, June 23-28, 2013}, pages 735--742,
  2013.

\bibitem{DBLP:journals/compsys/Mozer89}
Michael~C. Mozer.
\newblock A focused backpropagation algorithm for temporal pattern recognition.
\newblock {\em Complex Systems}, 3(4), 1989.

\bibitem{munkres1957algorithms}
James Munkres.
\newblock Algorithms for the assignment and transportation problems.
\newblock {\em Journal of the society for industrial and applied mathematics},
  5(1):32--38, 1957.

\bibitem{DBLP:conf/conll/NallapatiZSGX16}
Ramesh Nallapati, Bowen Zhou, C{\'{\i}}cero~Nogueira dos Santos, {\c{C}}aglar
  G{\"{u}}l{\c{c}}ehre, and Bing Xiang.
\newblock Abstractive text summarization using sequence-to-sequence rnns and
  beyond.
\newblock In {\em Proceedings of the 20th {SIGNLL} Conference on Computational
  Natural Language Learning, CoNLL 2016, Berlin, Germany, August 11-12, 2016},
  pages 280--290, 2016.

\bibitem{DBLP:journals/corr/RanzatoCAZ15}
Marc'Aurelio Ranzato, Sumit Chopra, Michael Auli, and Wojciech Zaremba.
\newblock Sequence level training with recurrent neural networks.
\newblock In {\em 4th International Conference on Learning Representations,
  {ICLR} 2016, San Juan, Puerto Rico, May 2-4, 2016, Conference Track
  Proceedings}, 2016.

\bibitem{DBLP:journals/pami/RenHG017}
Shaoqing Ren, Kaiming He, Ross~B. Girshick, and Jian Sun.
\newblock Faster {R-CNN:} towards real-time object detection with region
  proposal networks.
\newblock {\em {IEEE} Trans. Pattern Anal. Mach. Intell.}, 39(6):1137--1149,
  2017.

\bibitem{DBLP:journals/ijcv/RussakovskyDSKS15}
Olga Russakovsky, Jia Deng, Hao Su, Jonathan Krause, Sanjeev Satheesh, Sean Ma,
  Zhiheng Huang, Andrej Karpathy, Aditya Khosla, Michael~S. Bernstein,
  Alexander~C. Berg, and Fei{-}Fei Li.
\newblock Imagenet large scale visual recognition challenge.
\newblock {\em International Journal of Computer Vision}, 115(3):211--252,
  2015.

\bibitem{DBLP:conf/iccv/SadeghianAS17}
Amir Sadeghian, Alexandre Alahi, and Silvio Savarese.
\newblock Tracking the untrackable: Learning to track multiple cues with
  long-term dependencies.
\newblock In {\em {IEEE} International Conference on Computer Vision, {ICCV}
  2017, Venice, Italy, October 22-29, 2017}, pages 300--311, 2017.

\bibitem{DBLP:journals/corr/abs-1806-04936}
Stanislau Semeniuta, Aliaksei Severyn, and Sylvain Gelly.
\newblock On accurate evaluation of gans for language generation.
\newblock {\em CoRR}, abs/1806.04936, 2018.

\bibitem{DBLP:journals/corr/abs-1808-01562}
Han Shen, Lichao Huang, Chang Huang, and Wei Xu.
\newblock Tracklet association tracker: An end-to-end learning-based
  association approach for multi-object tracking.
\newblock {\em CoRR}, abs/1808.01562, 2018.

\bibitem{DBLP:conf/cvpr/SonBCH17}
Jeany Son, Mooyeol Baek, Minsu Cho, and Bohyung Han.
\newblock Multi-object tracking with quadruplet convolutional neural networks.
\newblock In {\em 2017 {IEEE} Conference on Computer Vision and Pattern
  Recognition, {CVPR} 2017, Honolulu, HI, USA, July 21-26, 2017}, pages
  3786--3795, 2017.

\bibitem{DBLP:conf/cvpr/TangAAS17}
Siyu Tang, Mykhaylo Andriluka, Bjoern Andres, and Bernt Schiele.
\newblock Multiple people tracking by lifted multicut and person
  re-identification.
\newblock In {\em 2017 {IEEE} Conference on Computer Vision and Pattern
  Recognition, {CVPR} 2017, Honolulu, HI, USA, July 21-26, 2017}, pages
  3701--3710, 2017.

\bibitem{DBLP:conf/naacl/TevetHSB19}
Guy Tevet, Gavriel Habib, Vered Shwartz, and Jonathan Berant.
\newblock Evaluating text gans as language models.
\newblock In {\em Proceedings of the 2019 Conference of the North American
  Chapter of the Association for Computational Linguistics: Human Language
  Technologies, {NAACL-HLT} 2019, Minneapolis, MN, USA, June 2-7, 2019, Volume
  1 (Long and Short Papers)}, pages 2241--2247, 2019.

\bibitem{DBLP:conf/nips/VaswaniSPUJGKP17}
Ashish Vaswani, Noam Shazeer, Niki Parmar, Jakob Uszkoreit, Llion Jones,
  Aidan~N. Gomez, Lukasz Kaiser, and Illia Polosukhin.
\newblock Attention is all you need.
\newblock In {\em Advances in Neural Information Processing Systems 30: Annual
  Conference on Neural Information Processing Systems 2017, 4-9 December 2017,
  Long Beach, CA, {USA}}, pages 5998--6008, 2017.

\bibitem{DBLP:conf/emnlp/WisemanR16}
Sam Wiseman and Alexander~M. Rush.
\newblock Sequence-to-sequence learning as beam-search optimization.
\newblock In {\em Proceedings of the 2016 Conference on Empirical Methods in
  Natural Language Processing, {EMNLP} 2016, Austin, Texas, USA, November 1-4,
  2016}, pages 1296--1306, 2016.

\bibitem{DBLP:conf/icip/WojkeBP17}
Nicolai Wojke, Alex Bewley, and Dietrich Paulus.
\newblock Simple online and realtime tracking with a deep association metric.
\newblock In {\em 2017 {IEEE} International Conference on Image Processing,
  {ICIP} 2017, Beijing, China, September 17-20, 2017}, pages 3645--3649, 2017.

\bibitem{DBLP:journals/corr/WuSCLNMKCGMKSJL16}
Yonghui Wu, Mike Schuster, Zhifeng Chen, Quoc~V. Le, Mohammad Norouzi, Wolfgang
  Macherey, Maxim Krikun, Yuan Cao, Qin Gao, Klaus Macherey, Jeff Klingner,
  Apurva Shah, Melvin Johnson, Xiaobing Liu, Lukasz Kaiser, Stephan Gouws,
  Yoshikiyo Kato, Taku Kudo, Hideto Kazawa, Keith Stevens, George Kurian,
  Nishant Patil, Wei Wang, Cliff Young, Jason Smith, Jason Riesa, Alex Rudnick,
  Oriol Vinyals, Greg Corrado, Macduff Hughes, and Jeffrey Dean.
\newblock Google's neural machine translation system: Bridging the gap between
  human and machine translation.
\newblock {\em CoRR}, abs/1609.08144, 2016.

\bibitem{DBLP:journals/corr/abs-1904-11489}
Jiarui Xu, Yue Cao, Zheng Zhang, and Han Hu.
\newblock Spatial-temporal relation networks for multi-object tracking.
\newblock {\em CoRR}, abs/1904.11489, 2019.

\bibitem{DBLP:conf/cvpr/YangCL16}
Fan Yang, Wongun Choi, and Yuanqing Lin.
\newblock Exploit all the layers: Fast and accurate {CNN} object detector with
  scale dependent pooling and cascaded rejection classifiers.
\newblock In {\em 2016 {IEEE} Conference on Computer Vision and Pattern
  Recognition, {CVPR} 2016, Las Vegas, NV, USA, June 27-30, 2016}, pages
  2129--2137, 2016.

\end{thebibliography}
}

\end{document}